# Contextual Unsupervised Outlier Detection in Sequences

Mohamed A. Zahran
Department of Computer Science,
Purdue University
West Lafayette, IN
mzahran@purdue.edu

Leonardo Teixeira
Department of Computer Science,
Purdue University
West Lafayette, IN
lteixeir@purdue.edu

Vinayak Rao
Department of Statistics,
Purdue University
West Lafayette, IN
varao@purdue.edu

Bruno Ribeiro
Department of Computer Science,
Purdue University
West Lafayette, IN
ribeiro@cs.purdue.edu

## ABSTRACT
This work proposes an unsupervised learning framework for trajectory (sequence) outlier detection that combines ranking tests with user sequence models. The overall framework identifies sequence outliers at a desired false positive rate (FPR), in an otherwise parameter-free manner. We evaluate our methodology on a collection of real and simulated datasets based on user actions at the websites last.fm and msnbc.com, where we know ground truth, and demonstrate improved accuracy over existing approaches. We also apply our approach to a large real-world dataset of Pinterest and Facebook users, where we find that users tend to reshare Pinterest posts of Facebook friends significantly more than other types of users, pointing to a potential influence of Facebook friendship on sharing behavior on Pinterest.

## 1 INTRODUCTION

In this paper, we consider the task of detecting outliers in a sequence of actions or items $\mathbf{x} = (x_1, \ldots, x_n)$. Traditional approaches to such tasks (e.g. credit-card fraud detection) require **labeled data**: the sequence $\mathbf{x}$ has outliers identified by an $n$-bit vector $\mathbf{y} \in \{0, 1\}^n$. Such labels in real-world sequences are expensive, as outliers can be rare and sequences can be long. Consequently, there is a need for unsupervised algorithms with **unlabeled data** that can find outliers without explicit labels. Here, outliers are actions that deviate from the 'normal', defined according to some statistical model. While there are a multitude of methods for surpervised outlier detection, unsupervised approaches have a large number of variations that are still open problems. We consider one such variation, which we describe as *outlier-detection-within-a-context* .

We extend Hawkins's definition of outliers [18] to *contextual outliers* in sequence data as follows (extension in italics) : "A *contextual* outlier is an observation which, *despite being common in the dataset*, deviates so much from the *context in which it appears* as to arouse suspicions that it was generated by a different mechanism." The identification and localization of outliers in sequences has a long history [17]: from distance-based models that rely on clustering [38] (e.g., $k$-nearest neighbors), such as normalized longest common subsequence [8], edit distance [9, 30], and compression-based dissimilarity[21], to more sophisticated statistical models such as hidden Markov models (HMMs) [22, 24, 39]. Keorgh et al. [21] proposes a frequency-based approach to determine the surprise level of comparison units, using information-theoretic measures.

As stated in our definition of outliers, we want sequence outliers to include not just rare events, but also *common occurrences that are rare within their context.* This is not properly handled by existing unsupervised methods. As an example, a user who listens to heavy metal frequently can have a sequence of songs: "Metallica→Megadeth →Lady Gaga→ Slayer→ Submersed". The outlier here is "Lady Gaga". However, "Lady Gaga" is so frequent among the users in the dataset, that a simple sequence model might flag the much less popular metal band "Submersed" as the outlier, despite the fact that it matches the context better than Lady Gaga.

One of the key challenges of outlier detection-within-a-context is that of multiple dependent hypothesis testing, where testing the null hypothesis $H_0^{(u,i)}$ (that Event $x_i$ of user $u$ is not an outlier) depends on the hypotheses $H_0^{(u,i+k)}, k \in \{\ldots, -1, 1, \ldots\}$ (that Event $x_{i+k}$ is not an outlier.) Thus, outlier detection-within-a-context should account for multiple dependent hypotheses. Only in recent years have contextual sequence models have become scalable, while steadily supplanting more traditional methods. Our work brings this philosophy to the problem of outlier detection-within-a-context.

### Challenges and Contributions
Bayesian sequence models [13, 14] and deep neural network sequence models (RNNs [25], LSTMs [19], GRUs [10]) allow us to learn the context of a sequence. However, directly using these methods for outlier detection is undesirable as they cannot automatically deal with challenges of multiple dependent hypotheses testing. Specifically one must control the False Positive Rate (FPR) of the outlier detection method.

*Contributions:* This paper proposes a parameter-free approach to detect contextual outliers in sequences of actions with a hybrid method that mixes Monte Carlo tests with rank tests. We take a computational approach, freeing hypothesis-testing from flawed assumptions and approximations. Our framework automatically adjusts its hyperparameters to meet a desired False Positive Rate

(FPR). This framework can be directly attached to any probabilistic model that explicitly models the context of a sequence. Our sampling approach significantly reduces computational complexity of testing for outliers in a length-$n$ sequence from $O(|\Omega|n)$ to $O(n)$, where $|\Omega|$ is the vocabulary size.

A second contribution, of independent interest, is a Bayesian personalized and interpretable sequence model we call *SwitchingFlow*. We show that *SwitchingFlow* outperforms standard Recurrent Neural Networks and Skipgram (word2vec) models for the task of outlier detection in our datasets. *SwitchingFlow* also outperforms LSTMs in most, but not all tasks.

*Reproducibility.* Our code will be made available online. All the data used in our evaluation is public domain.

## 2 RELATED WORK

Anomaly detection has a variety of applications in data analysis and data mining, including intrusion detection [3, 12, 23, 31, 33, 34], UNIX command and computer execution logs [28, 31, 32], human activity signatures [29], and fraud detection [4].

Sequeira and Zaki [34] proposed an intrusion detection system called ADMIT, using clustering of sequences of actions to identify masqueraders. Eskin et al. [12] map network trace data into a feature space and detect anomalies using SVM, k-NN, and cluster-based estimation techniques. Schonlau et al. [31, 32] use computer execution logs to detect masquerading behavior using Multi-step Markov models. Scott [33] proposed a Bayesian hierarchical approach to mine various levels of network traffic.

Cao et al. [6] propose a graph based fraud detection system for transactions called HitFraud. They model a database of inter-transactions dependency into a network, then use it to perform a collective fraud detection by monitoring the correlated and fast evolving fraudulent behaviors. Rashidi et al. [29] use an HMM approach for human activity mining and tracking. The majority of these approaches convert time sequences of actions into clusters to reduce the complexity, at the cost of some information loss.

Toledano et al. [36] present Anodot, a commercial system for anomaly detection in continuous time series with a special focus on financial data. The system works by choosing an appropriate model (e.g. parametric, non-parametric), then deals with seasonality detection and finally adapt system parameters according to the data. Zhou et al. [41] propose robust deep autoencoder which combines robust PCA and deep autoencoders. It splits data into two parts, one can be reconstructed by autoencoders and the second part is the noise (outliers) in the data.

Chandola et. al [7] survey anomaly detection on sequences, classifying them into: kernel-based, window-based and HMM-based techniques. Kernel-based techniques rely on calculating a similarity measure (for example, the longest common subsequence) between a test sequence and the training sets of normal and anomalous sequences. Window-based techniques divides an input sequence into overlapping subsequences of fixed length, assign an anomaly score for each subsequence, finally aggregating the scores of these subsequences to obtain an anomaly score.

The goal of most sequence anomaly techniques is, however, to classify the whole sequence rather than *localizing* the anomalous action in an existing sequence. Sequence models such as Hidden Markov Models (HMMs) and Recurrent Neural Networks (RNNs) are better suited for anomalous action localization, and we use these as either baselines or as null models within our framework. HMM-based techniques begin by learning an HMM model over the training data, then use the model to calculate the anomaly score for each test sequence. The negative log likelihood of the test sequence given by the model is considered to be its anomaly score. Warrender et. al. [37] used HMM for flagging outliers in a given sequences. Using the underlying transitions of HMM, they set thresholds for normal transitions and output probabilities. Then, emissions below these thresholds are flagged as abnormal.

## 3 PROBLEM DEFINITION

In what follows we use uppercase letters to describe stochastic processes and random variables, and lower case letters $x$ to describe their realizations. For user $u$, $x^{(u)}$ is an observation of a random sequence of actions $X^{(u)}$. We refer to the $i$-th action of $x^{(u)}$ as $x_i^{(u)}$.

*Definition 3.1.* The anomaly location problem under null model $\mathcal{M}_0^{(u)}$ of user $u$ tests against the null hypothesis $H_0^{(u,j)}$ that sequence $X^{(u)}$ of user $u$ has no outlier at the $j$-th action. The space $\Omega^{T^{(u)}}$ of all possible length-$T^{(u)}$ action sequences of user $u$ is divided into two disjoint sets, the *critical region, or region of rejection* $R_j^{(u)}$, where action $j$ is an outlier, and its complement, the region of acceptance $A_j^{(u)}$ (thus $R_j^{(u)} \cup A_j^{(u)} = \Omega^{T^{(u)}}$). The null hypothesis $H_0^{(u,j)}$ is rejected with probability $\beta \in (0, 1)$ if

$$P[X^{(u)} \in R_j^{(u)} | \mathcal{M}_0^{(u)}] > \beta \quad (1)$$

Definition 3.1 above gives guidelines to search for outliers: If the sequence of actions of user $u$, $x^{(u)}$, belongs to the rejection region $R_j^{(u)}$ with probability at least $\beta$ in the null model $\mathcal{M}_0^{(u)}$, then action $X_j^{(u)}$ is an outlier. The focus of our paper is about $R_j^{(u)}$ and $\beta$.

## 4 PROPOSED FRAMEWORK

The centerpiece of our method is the definition of the rejection region $R_j^{(u)}$ and a method to adjust $\beta$ in order to achieve a given FPR. For each user, we will be testing multiple hypotheses $H_0^{(u,j)}$, for $j = 1, \ldots, T^{(u)}$. Thus, the rejection probability $\beta$ (eq. (1)) must take into account the family-wise error, i.e., the total number of hypotheses $\{H_0^{(u,j)}\}_{\forall u,j}$ based on a false positive rate (FPR) $\alpha \in [0, 1]$ we are willing to tolerate. Because of the dependencies between the hypotheses in $\{H_0^{(u,j)}\}_{\forall u,j}$ one needs to properly scale $\beta$ to meet our maximum FPR of $\alpha$. Rather than using family-wise correction methods such as Bonferoni [11] and Hommel [20] for the scaling of $\beta$, we address the issue through a combination of a nonparametric rank test and a Monte Carlo method. We first present the basic outlier detection algorithm in Section 4.1 and the complete framework in Section 4.2. We introduce the existing null sequence models in Section 4.3. Finally, Section 4.4 introduces a new personalized sequence model, SwitchingFlow, which is also incorporated in our framework as a null model.

## 4.1 Basic Outlier Detection Algorithm

Algorithm 1 describes our basic outlier detection algorithm. The inputs to Algorithm 1 are: a sequence of actions $x^{(u)}$ of user $u$, the list of all possible actions $\Omega$, a null user model $\mathcal{M}_0^{(u)}$, and the rejection probability $\beta > 0$. $\beta$ helps define the rejection region of our algorithm; later we will eliminate this, replacing it with an FPR parameter $\alpha \in [0, 1]$. We will sometimes refer to the rejection region $R_j^{(u)}$ of user $u$'s $j$-th action as $R_j^{(u)}(\beta)$. The algorithm returns decisions (outlier or normal) for all actions in $x^{(u)}$.

The rejection region $R_j^{(u)}(\beta)$ is based on an empirical rank test outlined below. Consider user $u$'s sequence: $x^{(u)} = (x_1^{(u)}, \ldots, x_{T^{(u)}}^{(u)})$. We replace the $j$-th action $x_j^{(u)}$ with action $a \in \Omega$, obtaining a new sequence $x'^{(u)}(j, a)$. We calculate scores under the null model for each of these new sequences, $score_{j,a} = P[x'^{(u)}(j, a)|\mathcal{M}_0^{(u)}]/\hat{\gamma}_a$, where $\hat{\gamma}_a > 0$ is to reduce the probability of declaring $a$ an outlier action only because it is rare (and not because it does not fit the surrounding context). We set $\hat{\gamma}_a$ as the Laplace smoothed fraction of times action $a$ appears in the data, $\hat{\gamma}_a = (\epsilon + \#a)/(N\epsilon + \sum_{a' \in \Omega} \#a')$, with $\#a$ the number of times action $a \in \Omega$ is observed in the data (considering all users) and $\epsilon > 0$ a smoothing parameter.

Finally, as Algorithm 1 shows, we rank these new sequences $\{x'^{(u)}(i, a)\}_{a \in \Omega}$ in ascending order of their scores. The rejection region is defined by the $k = \lceil \beta \cdot |\Omega| \rceil$ lowest score sequences.

Ranking all actions $\{x'^{(u)}(i, a)\}_{a \in \Omega}$ is expensive for large $\Omega$. Below, we show how sampling without replacement can reduce this complexity, while having strong accuracy guarantees.

THEOREM 4.1 (ACTION SAMPLING). *Let $\Omega'$ be a set of actions sampled uniformly without replacement from all actions $\Omega$. Let $\delta$ be the difference in p-value (Algorithm 1) calculated using $\Omega$ and $\Omega'$ for an action $x_i^{(u)} \in x^{(u)}$. Then, for $\epsilon \geq 0$,*

$$P(\delta \geq \epsilon) \leq exp(-2\epsilon^2 |\Omega'|). \quad (2)$$

Note that eq. (2) does not depend on the total number of actions $|\Omega|$. Rather, the accuracy of our sampling depends only on how many actions we sample in $\Omega'$, independent of the size of $\Omega$. The independence from $|\Omega|$ appears because we are just trying to estimate the percentage of actions that lie below the user's chosen action. The following corollary further explores this property.

COROLLARY 4.2. *Let $|\Omega'| = \min(T, |\Omega|)$, with $\Omega'$ uniformly sampled without replacement from $\Omega$ and $T > 1$ constant. Then, for $\epsilon \geq 0$, $P(\delta \geq \epsilon) \leq \exp(-2T\epsilon^2)$.*

Corollary 4.2 is a direct application of Theorem 2, showing that the complexity of the algorithm with action sampling (Algorithm 1) goes from $O(|x^{(u)}||\Omega|)$ of the original algorithm to $O(|x^{(u)}|)$, a substantial speed-up for large vocabulary sizes. A good choice for $T$ is $T = 500$, which for $\epsilon = 0.05$ yields $P(\delta \geq 0.05) < 0.09$. In our experiments we show that sampling $T = 500$ actions gives similar results as using all actions in $\Omega$.

*Rejection region justification.* Monte Carlo statistical tests [1, 2] are designed to test whether the observed value of a real-valued statistic is "unreasonably large/small" according to a null model, and rely on a natural ordering of actions. Our strategy allows us to get around this absence of ordering with a rank test [35, Chapter 5]: we rank the observations from least likely to the most likely, and then define outliers to be the top $k$ observations, where $k = \lceil \beta \cdot |\Omega| \rceil$ and $\beta$ is a parameter of the algorithm.

---

**Algorithm 1** outlierDetection($x^{(u)}$, $\Omega$, $\mathcal{M}_0^{(u)}$, $\beta$, T)

$decisions \leftarrow$ empty list of length equals to $|x^{(u)}|$
**for** j = 1 : $|x^{(u)}|$ **do**
  $scores \leftarrow []$
  **for** a ∈ SampleWithoutReplacement($\Omega$,T) $\cup \{x^{(u)}[j]\}$ **do**
    {Samples $\min(T, |\Omega|)$ actions without replacement from $\Omega$, guaranteeing user action is in this set}
    $x'^{(u)} \leftarrow$ replace action at index $j$ in $x^{(u)}$ with $a$
    $scores[a] \leftarrow$ Use $\mathcal{M}_0^{(u)}$ to score $x'^{(u)}$
  **end for**
  $rank \leftarrow$ the rank of $x^{(u)}[j]$ in ascending order of $scores$
  $pvalue \leftarrow rank / \min(T, |\Omega|)$
  **if** $pvalue \leq \beta$ **then**
    $decisions[j] \leftarrow$ 'OUTLIER'
  **else**
    $decisions[j] \leftarrow$ 'NORMAL'
  **end if**
**end for**
return $decisions$

---

## 4.2 Framework Description: Training $\mathcal{M}_0^{(u)}$, Multiple Hypotheses, FPR

The basic algorithm presented in Section 4.1 is now augmented to take into account multiple hypotheses. Consider a desired False Positive Rate $\alpha \in [0, 1]$.

Figure 1 shows the layout of our framework as described below:
**(Phase A)** In Phase A, we learn the null model $\{\mathcal{M}_0^{(u)}\}_{u \in U}$ over a dataset of user action sequences. If possible (but not necessarily), we will have a separate training data that is used only to fit $\{\mathcal{M}_0^{(u)}\}_{u \in U}$, with outlier prediction kept separate from training. The null model should be regularized to avoid overfitting the training data, and to avoid fitting the outliers.
**(Phase B)** In Phase B, we will simulate data from the null models $\{\mathcal{M}_0^{(u)}\}_{\forall u}$. Clearly, any rejections of this simulated data are rejects of a true hypothesis (as the data comes from the null model). We use this step to calibrate $\beta$ to give a False Positive Rate (FPR) of at most $\alpha$, where $\alpha$ is the desired FPR. The value of $\beta$ is found via binary search, and is used in Phase C to define the rejection regions with the desired $\alpha$.
**(Phase C)** In Phase C, we apply Algorithm 1 to the sequences we want to detect outliers: $\{x^{(u)}\}_{u \in U}$. Note that we have already selected the rejection region parameter $\beta$ in Phase B. For the $i$-th action of user $u$, $x_i^{(u)}$, we declare $x_i^{(u)}$ to be an outlier action if $x^{(u)} \in R_i^{(u)}(\beta)$ as described in Section 4.1, i.e., the sequence $x^{(u)}$ belongs to the rejection region of the $i$-th action of user $u$.

## 4.3 The Null Models $\mathcal{M}_0^{(u)}$

Our approach can make use of any generative sequence model. Our empirical results use the following models: *SwitchingFlow*, a novel generative personalized modeling technique introduced in section 4.4, *Tribeflow* [13], a personalized trajectory model, as well as generic non-personalized models like Ngram Language Model [5], Recurrent neural network language model (RNNLM) [25], Long-Short Term Memory (LSTM) [19] and Word2vec (Skipgram) [27]. We also compare our approach against widely used anomaly detection baselines such as hidden Markov models (HMMs) and i.i.d. t-tests (*bag-of-actions*).

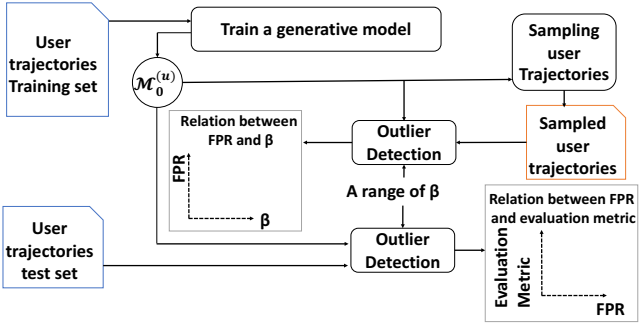

**Figure 1: High level layout of our framework for outlier detection.**

*4.3.1 Tribeflow [13].* Tribeflow is a general state-of-the-art personalized method that learns user behavior from a history window of $B$ actions to predict the next action, where $B$ is a hyper-parameter (the context window size). Personalization for each user is achieved through a per-user preference vector over a set of learned Markov chains called environments.

*4.3.2 Ngram Language Models.* Another fast way to model sequences is via Language Models (LM), treating the sequence of actions as words in a sentence. The Ngram LM is not personalized for each user, so that the model $\mathcal{M}_0$ is the same for all users. Ngram LM decompose the joint probability of a sequence of actions using the chain rule under Markov assumption that each action only depends on the previous $B$ actions.

*4.3.3 Recurrent Neural Networks (RNN).* Recurrent Neural Networks are popular techniques for sequence modeling. We follow the simpler procedure of Mikolov et al. [26] that encode sequences using an RNN. The network architecture has an input layer with a recurrent feed, one hidden layer and an output layer. The training procedure iterates over the sequences, one action at a time. The input layer receives the current action, the recurrent feed is the previously encoded context history, and the output layer is a softmax that generates a probability distribution over all possible actions. The RNN is trained using Back Propagation Through Time (BPTT) for up to $B$ steps, maximizing the likelihood of the data.

*4.3.4 Long Short Term Memory (LSTM).* While RNNs show great performance in sequence modeling, they cannot bridge long term dependencies. LSTM [19] solve this issue by using gating mechanism to regulate the flow of information between states. Each action in $\Omega$ is embedded into a learnable multidimentional vector, then the training procedure iterates over the sequences, one action at a time predicting the next action. The LSTM is trained with BPTT window greater than $B$ steps.

*4.3.5 Skip-gram [Word2vec] (W2V).* Proposed by Mikolov et al. [27], this is a widely used technique for embedding words in a multidimensional vector space. The idea is to use a running window of fixed size $B$ to scan over sequences of data. The embedding of each word in the sequence is optimized to predict all the other words in the same window using their vector embeddings. When training is over, each action's embedding can be used to calculate the probabilities of the sequences.

### 4.4 SwitchingFlow

In this section we describe our new personalized / contextual sequence model, *SwitchingFlow*. As described earlier, Tribeflow assumes a user performs a fixed number $B$ actions in one context or environment, then jumps to a new context to perform another $B$ actions, and so on. *SwitchingFlow* generalizes this by modeling the jump points with another process, so that environment switches do not necessarily have to be $B$ actions apart.

Assume $M$ different environments, each environment defining a different distribution over actions (which we model as a Markov Chain). We model the transition between environments themselves as a Markov Chain.

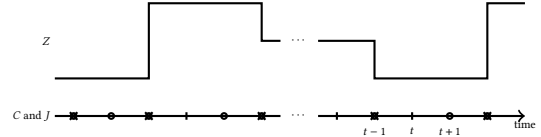

**Figure 2: Markov Chain of jump candidates and latent environment**

Figure 2 illustrates this latter Markov process for a particular user. For each discrete point $t \in \{1, \ldots, T_u\}$ in time, let $C_t$ be a Bernoulli random variable which indicates if this point is a candidate for a jump between environments (in the figure, $C_t = 1$ is denoted by a circle). We define $C_1 = 1$ and, for other points,

$$C_t \sim \text{Bernoulli}(q).$$

$q$ is an *uniformization* tuning parameter, chosen by grid search. For each candidate point, a decision is made to jump or not, according to the current environment $m$, with probability $\rho_m$. We follow this two-stage process for computational reasons, so that the forward-filtering backward-sampling scheme described later scales only with the number of candidate times, and not the *total* number of time points.

When a jump occurs, the new environment is chosen according to a user preference $P_u$. Let $J_t$ be a binary random variable indicating whether a jump occurred at time $t$ and $Z_t$ be the environment at time $t$. This defines $(J, Z \mid C, \rho, P_u)$ as a Markov Chain:

$$P(J_t = j \mid Z_{t-1} = m, C_t) = \begin{cases} \rho_m^j (1-\rho_m)^{1-j} & \text{if } C_t = 1 \\ \delta(j=0) & \text{if } C_t = 0 \end{cases}$$

$$P(Z_t = z \mid J_t = j, Z_{t-1} = m) = \begin{cases} P_u(z) & \text{if } j = 1 \\ \delta(z=m) & \text{if } j = 0 \end{cases}$$

We define $J_1 = 1$. In the figure, the jumps are denoted by the crossed circles. Given $J, Z$, the user trajectory $x = (x_1, \ldots, x_{T_u})$ defines a Markov Chain (or a set of Markov Chains, with each chain being associated with one environment):

$$P(x_t \mid x_{t-1}, Z_t, J_t, \mathcal{A}) = \mathcal{A}(Z_t, x_{t-1}, x_t)^{(1-J_t)} \frac{1}{N}^{J_t},$$

where the element $(z, i, j)$ of tensor $\mathcal{A}$ gives the transition probability between states $i \in \Omega$ and $j \in \Omega$ at environment $z \in \{1, \ldots, M\}$. The $1/N$ terms are because we uniformly pick a new state when we enter a new environment.

To complete the model, we specify prior distributions for $\mathcal{A}$ and $\rho$, with $z \in \{1, \ldots, M\}, i \in \Omega$:

$$\mathcal{A}(z, i, \cdot) \sim \text{Dir}(a)$$
$$\rho_z \sim \text{Beta}(r, s), \quad r, s \in \mathbb{R}^+.$$

*4.4.1 Gibbs Sampler.* To construct a Gibbs sampler for this model, we must be able to sample from the full conditional distributions. The conditional probability of time $t$ being a jump candidate is given by:

$$P(C_t \mid C^{-t}, J, Z, \mathcal{A}, \rho, x) \propto \begin{cases} \left(q\left(1 - \rho_{Z_{t-1}}\right)\right)^{C_t} (1-q)^{1-C_t} & \text{if } J_t = 0, \\ q\rho_{Z_{t-1}}\delta(C_t = 1) & \text{if } J_t = 1 \end{cases}$$

which is summarized in Table 1.

|  | $C_t = 0 \mid J_t$ | $C_t = 1 \mid J_t$ |
|---|---|---|
| $J_t = 0$ | $\frac{1-q}{1-q\rho_{Z_{t-1}}}$ | $\frac{q-q\rho_{Z_{t-1}}}{1-q\rho_{Z_{t-1}}}$ |
| $J_t = 1$ | 0 | 1 |

Table 1: Conditional probabilities for sampling of $C_t$

The conditional posterior for $\mathcal{A}(z, i, \cdot)$ is the updated Dirichlet:

$$\mathcal{A}(z, i, \cdot) \sim \text{Dir}\left(a + \#_{i \to j}^z\right), \quad (3)$$

in which $\#_{i \to j}^z$ is the count of transitions from state $i$ to state $j$ while on environment $z$. We sample the jumps and environments, $J$ and $Z$, using the Forward Filtering Backward Sampling (FFBS) algorithm [15]. Given $C$, we can separate $J, Z$ into two sets: $J^C, Z^C$ which are the jump and the environment variables for the candidate points ($C_t = 1$) and $J^{\neg C}, Z^{\neg C}$, for the points which are not candidate ($C_t = 0$). Moreover, the noncandidate variables $J^{\neg C}, Z^{\neg C}$ are completely defined by $J^C$ and $Z^C$, so we need only to sample $J^C, Z^C$. Since we must focus only on the candidate jump points, FFBS scales only with the number of these points, which is typically much less than the total number of times points. We denote the candidate jump points by $J_i$ and $Z_i$ ($i = 1, \cdots, T_C$ are the indices of these points). We also denote the vector of observations between jump candidates $i$ and $i+1$ as $x_{(i)} = \{x_t : t_i \le t < t_{i+1}\}$ ($t_i$ is the index of the $i$-th jump candidate). The forward filtering step involves computing the forward probabilities $\alpha_{jz}(i) = P(x_{(1)}, \cdots, x_{(i)}, J_i = j, Z_i = z \mid \mathcal{A}, C, P_u, \rho)$:

$$\alpha_{jz}(1) = P_u(z) \frac{1}{N} A_e^*(x_{(1)})$$
$$\alpha_{jz}(i) = \sum_{h=0}^{1} \sum_{m=1}^{M} \alpha_{hm}(i-1)\gamma_{hm}^{jz}(i)$$

The function $\gamma_{hm}^{jz}(i) = P(x_{(i)}, J_i = j, Z_i = z \mid x_{(1)}, \cdots, x_{(i-1)}, J_{i-1} = h, Z_{i-1} = m, \mathcal{A}, C, P_u, \rho)$ is computed as:

$$\gamma_{hm}^{jz}(i) = \begin{cases} \rho_m P_u(z) \frac{1}{N} A_z^*(x_{(i)}) & \text{if } j = 1 \\ (1 - \rho_m)\delta(z = m) A_z^*(\{x_{(i-1)}^{-1}\} \cup x_{(i)}) & \text{if } j = 0 \end{cases}$$

in which $x_{i-1}^{-1}$ is the last variable of the vector $x_{(i-1)}$ and

$$A_z^*(x) = \prod_{t=t_s+1}^{t_e} \mathcal{A}(z, x_{t-1}, x_t),$$

in which $t_s$ and $t_e$ index the first and last variables of the set $x$.

With this, the backward sampling step can be performed as:

$$P(J_i = j, Z_i = z \mid J_{i+1} = h, \cdots, J_{T_C}, Z_{i+1} = m, \cdots, Z_{T_C},$$
$$x_{(1)}, \cdots, x_{(T_C)}, \mathcal{A}, C, P_u, \rho) \propto \alpha_{jz}(i)\gamma_{jz}^{hm}(i+1).$$

Putting all these steps together gives the *SwitchingFlow* Gibbs sampler. The final result is a Bayesian personalized sequence model. Our code will be made available as open source.

## 5 RESULTS AND EVALUATION

In this section we evaluate the performance of our approach in three settings: (a) *Simulated data (without outliers) with injected simulated outliers for precision and recall tests*: Last.fm [13] is an online music server where users advertise the songs they are currently listening to. We fit a sequence model to this data, from which we simulate outlier-free sequences. To these sequences, we then inject artificial outliers, allowing us to test precision and recall of different methods in recovering outliers under controlled conditions. (b) *Real data, injected simulated outliers for recall test:* With user browsing logs from msnbc.com, we injected artificial outliers, and evaluated the ability of different methods to correctly identify all injected outliers. The original (real) data may or may not have outliers of its own, thus, we do cannot test precision in this setting. (c) *Real data, real outliers:* We used a dataset from Pinterest, consisting of pins, repins, and likes, carrying out a qualitative and quantitative analysis of unsupervised outliers flagged by our method. In all cases, test data is restricted to sequences having at least 9 actions. We describe out three datasets next.

### 5.1 Last.Fm Dataset

This dataset records user activity on last.fm[1], an online music streaming service. Measurements include user IDs, artists, songs, and the time stamp when a user plays a song. We treat the data as the sequence of artists that a user has listened to, with artist names forming the actions. Table 2 shows the statistics of the dataset, and we restrict the data to the top 5000 most frequently played artists. Since we have no ground truth here, we use this data to artificially create outliers. In particular, we first cluster users into 3 groups based on their 200 most frequent artists. We produce these clusters using k-means, and for each cluster, we train a different 10-state HMM. These three HMMs are then sampled to generate artificial sequences, which are then injected with artificial outliers. The injection process randomly selects artists in a sequence, and replaces them with random samples from the most frequent artists

---
[1]http://www.dtic.upf.edu/~ocelma/MusicRecommendationDataset/lastfm-1K.html

Table 2: LastFm and MSNBC datasets statistics.

| Item | Last.Fm | MSNBC |
|---|---|---|
| Number of users | 982 | 3203 |
| Number all possible actions $|\Omega|$ | 5000 | 17 |
| Number of actions in testset | 24550 | 32030 |
| % of injected outliers | 8% | 11% |

Table 3: Pinterest dataset statistics for sequences greater than history size ($B$).

| Item | Count |
|---|---|
| Number of users | 372k |
| Number of Pins, Repins | 3.9M |
| Number of Likes | 1.58M |
| Number of Categories | 369 |
| % of likes where owner&*likee* are Facebook friends | 0.0038% |

of a different cluster. The task here is to identify these injected outliers at a desired FPR.

### 5.2 MSNBC Dataset

This dataset [2] records sessions of page views for different users on msnbc.com. Each session is a sequence of web pages visited by a certain user. The web pages are categorized by topic (E.g. news, tech, weather etc). The session for each user is split into training and testing sequences with the test sequence the last $k$ web pages (we set k=10 in our experiments). Test sequences are assumed to be free of outliers and hence they are injected with artificial outliers. The injection process randomly selects web pages from the sequences and replaces them with different random web pages. The task again is to identify injected outliers at a given FPR.

### 5.3 Pinterest Dataset

The Pinterest dataset from Zhong et al. [40] records user activities (pins, repins, likes, timestamps, and Facebook friendship with other Pinterest users) on the website Pinterest. Each user has a *board* (an URL bookmark space). *Pins* refer to the websites a user pins to his/her board. *Repins* for user Alice are websites published by another user Bob, which Alice pinned to her board. *Likes* of user Alice are websites published by another user Bob which Alice liked but these do not appear on Alice's board. Because these website URLs tend to be unique, we aggregate them by topic, using the classification service K9 Blue-coat[3], getting 369 website categories, spanning 75% of all URLs in the dataset. The remaining 25% unclassified URLs were removed. Data statistics are shown in Table 3. We also have Facebook friendship links between the Pinterest users.

We assume that pinned and repinned URLs show the true interest of the users, as these appear on the user's board page. We train our null models on these pin and repin sequences. We then use these as null models in our framework, to detect outliers in user "likes". A Pinterest user Alice may *like* the pin or repin of another Pinterest user Bob. *Likes* do not appear on the user's board page, and may not indicate their true preferences of websites to visit.

As there is no ground truth in this dataset, we correlate the actions flagged as outliers by our method with whether or not the users are Facebook friends. Our motivation is what we call a *social like*: Alice's *like* of Bob's pin or repin is influenced by their friendship and not by Alice's true preferences. A dependence between *like* outliers and Facebook friendship indicates the presence of an anomaly (a *social like* or some other Facebook-related factor).

### 5.4 Evaluation Metrics for Pinterest Outliers

In the absence of ground truth, we propose two metrics to assess the quality of the outliers in the Pinterest dataset. Let $D$ be a random variable representing the framework's decision about a given *like*: this is either an outlier 'o' and normal 'n'. The like is given by user $u_x \in U$ to an item published by user $u_y \in U$. Let $R$ be a random variable representing a Facebook friendship link, true 't' or false 'f', between $u_x$ and $u_y$. We will say a method is good at flagging outliers if an outlier flagging $D = o$ means that users $u_x$ and $u_y$ are more likely than chance to have $R = t$.

**Frequentist Approach (Fisher's exact test)**: Here we use Fisher's test of independence to show that our framework's decision for a *like* is not independent from the friendship status between the user and the owner. Here, the null hypothesis $H_0$ is that $D$ and $R$ are independent. $H_0$ is rejected if the p-value of Fisher's test is less than a certain significance level ($\gamma$).

**Bayesian Approach**: While Fisher's test can tell if two variables are independent or not, its *p*-value does not indicate the direction of this dependence. For example, a model flagging all normal actions as outliers and all outliers as normal will make $D$ and $R$ dependent, but this in the opposite direction that we want. In our Bayesian hypothesis test formulation, we take into account the direction of dependency. Let $i_{dr}$ be the number of actions with $(D = d, R = r)$, $d \in \{o, n\}$ and $r \in \{t, f\}$. Now we calculate a Bayesian p-value, giving the probability that the outlier decision $D$ and the friendship status $R$ are independent:

$$1 - P[P(R = t|D = o) > P(R = t)]. \quad (4)$$

Let

$$Pr(t) = \frac{i_{ot} + i_{nt}}{i_{ot} + i_{nt} + i_{of} + i_{nf}}, \quad (5)$$
$$Pr(f) = 1 - Pr(t).$$

Assume $P(R = t)$ follows Beta distribution and $P(R = r|D = d)$ follows Dirichlet distribution, then

$$\begin{aligned} P(R = t) &\sim \text{Beta}(i_{ot} + i_{nt} + Pr(t), i_{of} + i_{nf} + Pr(f)) \\ P(D = d, R = r) &\sim \text{Dir}(i_{dr} + P(r) \times c) \end{aligned} \quad (6)$$

Where $c$ is a scaling constant. Form equations 5, 6 we can calculate the p-value in equation 4.

In our experiment, all models are trained over the pins, repins trajectories then employed on the trajectories of likes as a test set to detect outliers (social likes). For each model a figure is plotted between the desired FPR calculated from sampled data, and the p-values for Fisher and Bayesian tests measured on the Likes trajectories.

### 5.5 Baselines

Our empirical evaluation compares with two popular baselines: a frequency-based *Bag-of-actions* model, and a *Hidden Markov Model*.

---

[2]https://archive.ics.uci.edu/ml/datasets/MSNBC.com+Anonymous+Web+Data
[3]http://sitereview.bluecoat.com/sitereview.jsp

*Bag-of-Actions.* This approach assumes actions are i.i.d. samples from an unknown distribution. A given action is flagged as an outlier if its empirical probability is less than a certain threshold $\beta'$ (analogous to the rejection probability of our framework). To get an FPR for this method, we look at various such thresholds $\beta'$, computing the FPR rate for each threshold value using a null model that samples actions i.i.d. from the empirical distribution of actions.

*Hidden Markov Model (HMM).* Since each individual user may not have enough actions to train the HMM, we grouped sequences of all user actions to train a larger HMM across all users. For the Pinterest data with 369 categories, we observed that ten HMM states are enough properly describe all categories without underfitting or overfitting. Given a new sequence of actions $S = x_1, x_2, x_3, \ldots, x_j, \ldots, x_T$ we use the Viterbi algorithm to predicts the most likely sequence of hidden states $H_S = h_1, h_2, h_3, \ldots, h_j, \ldots, h_T$ that generated these actions. An action $x_j$ is flagged as an outlier if the emission probability $P(x_j|h_j)$ is less than a certain threshold (analogous to the rejection probability of our framework). To study the impact of thresholding on the outlier detection, the HMM detection method is applied to sampled sequences from the model itself which is assumed to be free of outliers and FPR is calculated accordingly.

### 5.6 Results

#### 5.6.1 Last.Fm Injected outliers.
Figure 3-(a) shows the injected outlier detection performance of all the null models within our framework, as well as the two baselines (bag-of-actions and HMM[4]). The figure shows the F1 score of detecting the injected outliers against the desired false positive rate $\alpha$. A good model will reach high F1 score with small FPR, after which the F1 score should decay slowly with increasing FPR. It is clear that SwitchingFlow with our framework significantly outperforms all other techniques in almost all the FPR scale [5]. We also used the HMM as a null model in our framework, this resulted in performance comparable to the baseline HMM. It worth noting that the baseline HMM outperforms some of the other null models (as Ngram LM) because the synthetic test data was sampled from 3 HMM's, giving the HMM some advantage. Figure 3-(b) also shows that the performance of null models in our framework with 500 sampled actions is remarkably similar to that using the full action set of 5000 actions.

*Qualitative Analysis of Last.Fm Outliers.* Figure 5 two representative examples of sequence outliers for a qualitative analysis for the outliers detected by our framework in Last.Fm experiment using SwitchingFlow at FPR = 0.05. **Note that we hide all genre information from our method to test whether it correctly produces outliers based on the learned context.**

In the sequence of Figure 5(a), the user is listening to the music genre *metal*. The flagged as outliers are dance, pop, indie and alternative. While these artists flagged as outliers have higher probabilities of occurrences in the data ("Artist prob") than other artists in the sequence, they do not match the user's context. For instance, the genre of "lost Horizon" is metal, and even though the probability of occurrence of "lost Horizon" is very low (4e-5), it is not flagged as

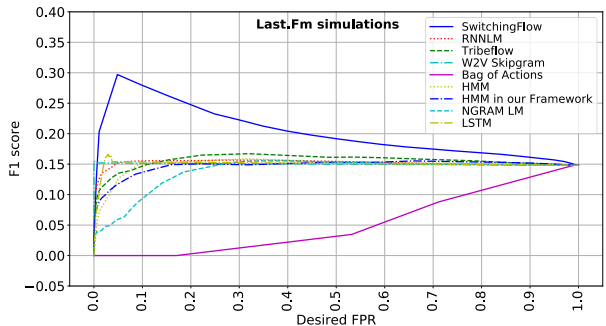

(a) LastFm simulations without action sampling($|\Omega'| = |\Omega| = 5000$)

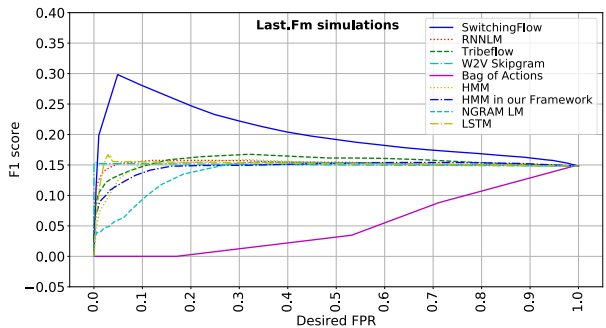

(b) LastFm simulations with action sampling($|\Omega'| = 500, |\Omega| = 5000$)

Figure 3: F1 score of outlier detection for all techniques on LastFm's injected data set against the desired false positive rate $\alpha$ with and without action sampling.

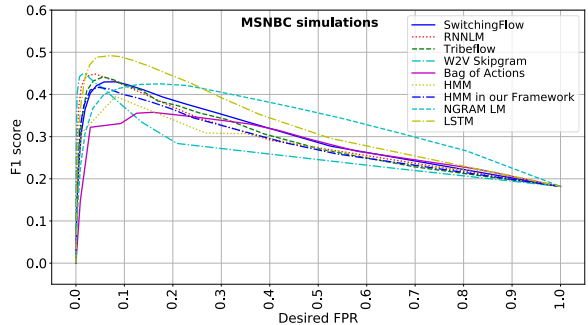

Figure 4: MSNBC outlier detection F1 measure against the desired false positive rate $\alpha$ for different null models.

outlier by our method as it belongs to the context of the sequence. Marilyn Manson is a popular artist (more popular than the outlier Jamiroquai) but he not flagged as an outlier as he better matches the context of the sequence.

In the sequence of Figure 5(b), the user is interested in j-pop and classical music. The flagged outliers are rock and chillout. The outliers have higher probabilities of occurrences in the data ("Artist prob") than other artists in the same sequence which are not flagged as outliers, but note that the artists not flagged as outliers match the context of the sequence. For instance, "Two-Mix"'s is j-pop, its

---
[4]The baseline HMM used for outlier detection is trained with 30 states.
[5]A qualitative analysis of outliers is included in the appendix

### (a) Sequence with *metal* context

| Artist name | Artist prob | p-value | Artist genre |
|---|---|---|---|
| Bolt Thrower | 6.03E-05 | 0.68 | death metal, metal |
| Dissection | 0.000119 | 0.358 | black metal, melodic black metal |
| Dio | 0.000339 | 0.125 | heavy metal, hard rock, metal |
| Skid Row | 0.000315 | 0.135 | hard rock, heavy metal |
| X | 0.000163 | 0.651 | punk rock, rock |
| Marilyn Manson | 0.001653 | 0.052 | Metal, gothic metal |
| Borknagar | 9.99E-05 | 0.41 | black metal, viking metal |
| The Duskfall | 7.22E-05 | 0.978 | Melodic Death Metal, death metal |
| Paradise Lost | 0.000732 | 0.569 | Gothic Metal, doom metal, metal |
| Diary Of Dreams | 0.000128 | 0.995 | darkwave, Gothic, industrial, electronic |
| Borknagar | 9.99E-05 | 0.425 | black metal, viking metal |
| Mercyful Fate | 0.000121 | 0.358 | heavy metal, black metal |
| Summoning | 0.000103 | 0.428 | black metal, Symphonic Black Metal |
| **_Jamiroquai_** | **_0.001101_** | **_0.02_** | **_funk, acid jazz, dance, pop_** |
| **_Bloc Party_** | **_0.002684_** | **_0.003_** | **_indie, alternative, british_** |
| Steve Vai | 0.000173 | 0.227 | instrumental rock, rock |
| Mors Principium Est | 0.000114 | 0.371 | Melodic Death Metal, death metal |
| Vibrasphere | 8.12E-05 | 0.823 | progressive trance, psychedelic |
| Deathstars | 0.000244 | 0.306 | industrial metal, Gothic Metal |
| Lost Horizon | 4.00E-05 | 0.922 | Power metal, Progressive metal, metal |
| Aesthetic Perfection | 9.38E-05 | 0.419 | dark electro, electro-industrial |
| Ayreon | 0.00022 | 0.204 | Progressive metal, Progressive rock, metal |
| Apulanta | 0.000141 | 0.326 | rock , punk |
| Turisas | 0.000115 | 0.391 | folk metal, viking metal, battle metal |
| Savatage | 0.000103 | 0.437 | heavy metal, Progressive metal, Power metal |

### (b) Sequence with *j-pop and classic* context

| Artist name | Artist prob | p-value | Artist genre |
|---|---|---|---|
| _Radiohead_ | _0.006113758_ | _0.0002_ | _alternative rock, rock_ |
| Megumi Hayashibara | 6.36E-05 | 0.602 | j-pop, japanese |
| George Frideric Handel | 0.00010984 | 0.3962 | Classical, Baroque, instrumental |
| Hello Saferide | 0.000140633 | 0.318 | pop, indie, female vocalists |
| Hellogoodbye | 0.000313979 | 0.1394 | indie, powerpop, pop |
| Orbital | 0.000521039 | 0.074 | electronic, techno, dance |
| **_Thievery Corporation_** | **_0.001173473_** | **_0.0228_** | **_chillout, trip-hop, electronic, lounge_** |
| Heitor Villa-Lobos | 4.79E-05 | 0.792 | Classical, contemporary classical |
| Uverworld | 0.000111511 | 0.3914 | J-rock, japanese, j-pop, anime |
| Phantom Planet | 0.000233813 | 0.186 | alternative, indie rock, alternative rock |
| Yui | 0.000101392 | 0.416 | j-pop, japanese, female vocalists, J-rock |
| Vnv Nation | 0.000476956 | 0.0874 | futurepop, electronic, synthpop |
| Two-Mix | 3.85E-05 | 0.9656 | j-pop, japanese, anime, electronic |
| Holly Golightly | 7.75E-05 | 0.5426 | female vocalists, Garage Rock, indie |
| My Rockets Up | 4.61E-05 | 0.8344 | indie rock, post-grunge, alternative |
| Chemistry | 0.000318993 | 0.1352 | j-pop, japanese, trance |
| varttina | 0.000155292 | 0.3494 | folk, Scandinavian folk, female vocalists |
| Johann Sebastian Bach | 0.000405712 | 0.1332 | Classical, baroque, instrumental |
| Hikaru Utada | 0.000189914 | 0.3524 | j-pop, japanese, female vocalists |
| High And Mighty Color | 0.000105274 | 0.6016 | J-rock, japanese, j-pop, female vocalists |
| Joseph Haydn | 5.61E-05 | 0.705 | Classical, baroque, composers |
| Asian Kung-Fu Generation | 0.000147633 | 0.3004 | J-rock, japanese, rock, j-pop |
| Underoath | 0.000226904 | 0.1904 | rock, post-hardcore, hardcore |
| Dangerdoom | 0.000190322 | 0.2322 | Hip-Hop, underground hip-hop |
| Absurd Minds | 0.000107734 | 0.4054 | futurepop, synthpop, electronic |

**Figure 5: Qualitative Analysis of the outliers detected in Last.Fm experiments. Each table represent a test sequences for a user. Artists are listed (top-down) according to their order in the test sequence. "Artist prop" is the probability of each artist calculated independently from the training data. "p-value" is the p-value calculated by our framework. "Artist genre" is a list of genre tags for each artist. Rows in bold are outliers as detected by our framework. Underlined rows are the true injected outliers.**

probability of occurrence is low (3e-5) but it's not flagged as outlier, as our method learns it matches the context of this sequence.

*5.6.2 MSNBC Injected outliers.* Figure 4 shows the outlier detection F1 measure against the desired FPR. For the baseline models we can see the bag of action model is the worst performing model. HMM in our framework outperforms baseline HMM. We should notice that most models are performing much better in the MSNBC dataset compared to Last.Fm because the number of actions $|\Omega|$ in MSNBC is much smaller that than of Last.Fm which made the job for these models a lot easier and that's why LSTM is the best performing model. As $|\Omega|$ decreases, the personalization aspect becomes less clear and so the difference between one user and another decreases which made a strong model as LSTM able to build a generic model that works for different users. SwitchingFlow on the other hand has an edge in modeling datasets with large $|\Omega|$ due to its way of modeling different latent environment and jumps between them.

*5.6.3 Social likes and outliers.* Figure 6 shows the results of the baseline models (bag of actions, and HMM) against all null models using our framework in detecting outliers (social likes) from the likes-sequences. Performance was measured by testing for independence between the framework's outlier decision $D$ and the Facebook friendship indicator $R$.

In order to assess the performance of a model based on these figures we consider the following criteria for a constant $0 < \gamma < 1$ (usually $\gamma = 0.05$). First: *How fast does a model reach small p-values for the Bayesian and Fisher test?* The best model will reach p-values $< \gamma$ with the minimum FPR. Second: *How does a model perform with increasing FPR?*

In an ideal null model, the relation between FPR and the p-values of Bayesian's and Fisher's test should have an "U" shape. Because, for small (large) values of FPR, very few (too many) outliers are flagged, which means the p-values are large due to the little evidence to relate $D$ and $R$. For FPR values between the two extremes, an ideal model would give outliers the lowest ranks in the sorted order of their possible replacements $\{x'^{(u)}(j,a)\}_{a \in \Omega}$, and thus with increasing FPR the model would flag only the true outliers which makes the p-values drop to refute the independence between $D$, $R$ giving a "U" shape. According to the above criteria, Figures 6(a-b) shows that most null models within our framework outperform all baselines without our framework: bag-of-actions (Figures 6(a)) and HMM (Figures 6(b)) without our framework. Both the bag-of-actions and the HMM model could not find any suitable FPR at which social likes are flagged as outliers which makes the p-values for both Fisher's and Bayesian's to be insignificant. Moreover, the Bayesian's p-value spikes while Fisher's p-value plumps for the same FPR range resulting in the *gap* shown in Figure 6(a). This shows that the model consistently flags true outliers as normal actions and vice versa.

Among the null models used in our framework, our proposed SwitchingFlow, Figure 6(d), outperforms all other null models. LSTM and RNN LM Figures 6(i),(f) are close second, which show that recurrent neural network models are effective in sequence modeling. Even though these models are trained on generic data they perform well in a personalized test. LSTM is superior to RNN LM which confirms that using long history improves sequence modeling. Tribeflow Figure 6(e) is ranked third, although the p-values for both tests quickly drop with small FPR, the model failed to maintain low p-values with increasing FPR as compared to other models.

To further understand the effectiveness of our framework, we used the baseline HMM as one of the null models within our framework for outlier detection. Using our framework (Figure 6-c), HMM was able to find a suitable rejection probability with a much lower FPR (compared to the baseline HMM) to have p-values for both Fisher's and Bayesian's test below the significance threshold, albeit the region is rather small.

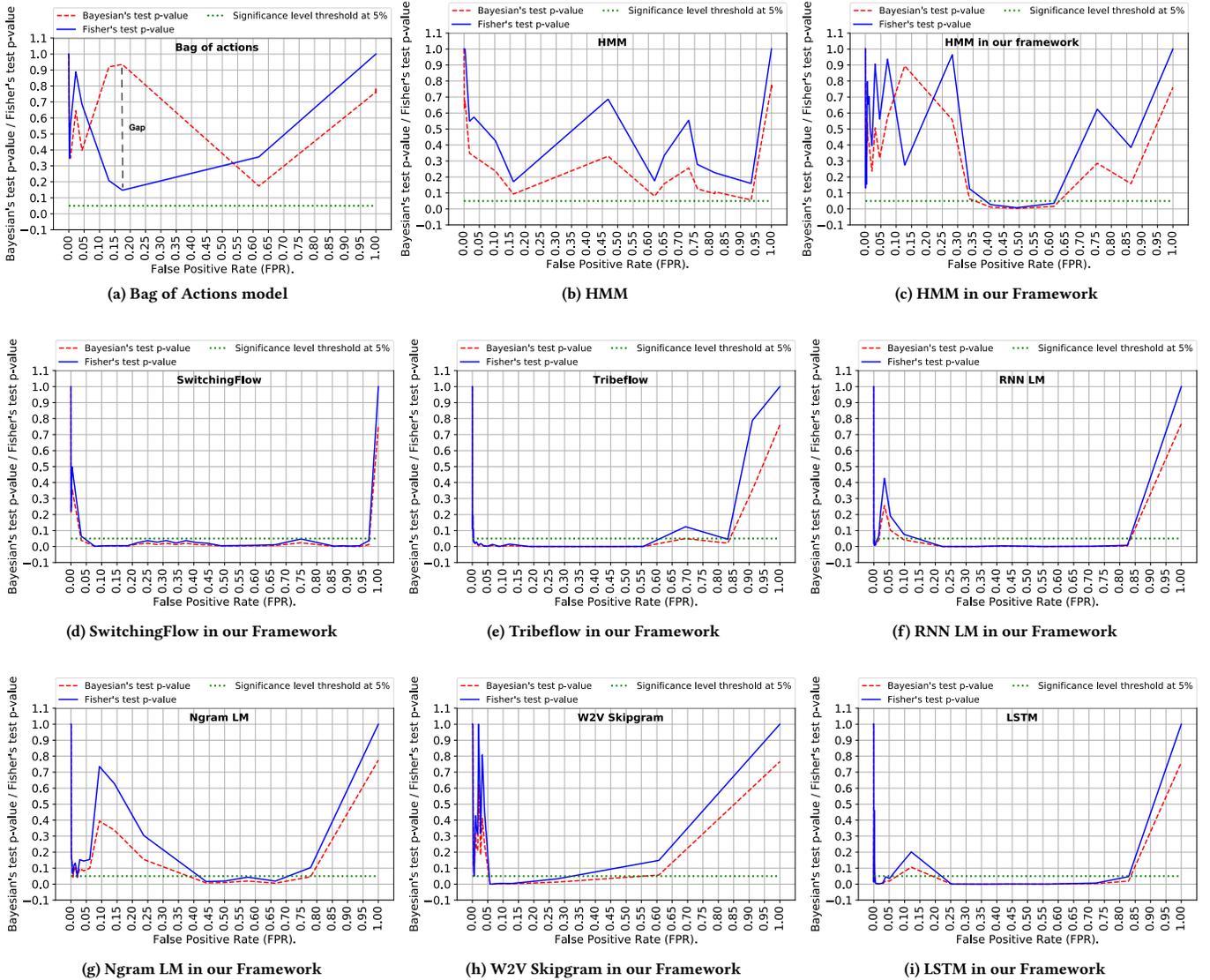

Figure 6: The relation between the desired FPR and the p-values of both Bayesian's and Fisher's tests for detecting outliers (social likes) in the likes sequence using all the null models $\mathcal{M}_0^{(u)}$ and the two baselines.

*Scalability with respect to vocabulary size (number of actions):* Figure 7 shows the execution time of different models in our framework per user sequence (25 actions long) as a function of the number of actions $|\Omega|$ with and without action sampling. The number of actions sampled $|\Omega'| = min(|\Omega|, 500)$. We note that SwitchingFlow, while accurate, is also the second slowest of all the methods. But, most importantly, sampling shows constant complexity and increasing speed up as $|\Omega|$ increases.

## 6 CONCLUSION AND FUTURE WORK

In this paper we presented a framework the uses non-parametric rank tests for outlier detection in sequences and showed how the method can be calibrated to meet a specific FPR. We also introduced SwitchingFlow: a Bayesian generative method to learn user sequences. We evaluated our outlier detection framework using three datasets. The first is simulated from Last.Fm music-listening data, with ground-truth (artificially injected) outliers. This shows our technique with SwitchingFlow to significantly outperforms all baseline methods. The second dataset is user browsing logs of msnbc.com with ground-truth (artificially injected) outliers. Which shows that our outlier detection framework outperforms HMM and i.i.d. outlier detection techniques. The third dataset comes from Pinterest and Facebook; while this dataset does not have ground-truth outliers, our method detects outliers that are highly correlated with friendship relations on Facebook (*social likes*), even though the null models are oblivious to such relationships. Finally, we showed that action sampling is a simple yet effective technique to significantly reduce the execution time of our framework.

# 7 APPENDIX
## 7.1 Proof of Theorem 4.1

For a sequence of actions: $S^{(u)} = x_1^{(u)} \ldots, x_i^{(u)}, \ldots, x_{T_u}^{(u)}$ and a set of actions $\Omega$. It's required to get the rank of $x_i^{(u)}$ among all of its possible replacements from $\Omega$. Let the set of actions with rank smaller than $x_i^{(u)}$ be $L$ and those of higher or equal rank be $R$. The true rank of $x_i^{(u)} = |L|$, which makes its true p-value ($pv$) $= \frac{|L|}{|\Omega|}$. Let $\Omega'$ be a smaller set of actions randomly sampled without replacement from $\Omega$. We can calculate the rank of $x_i^{(u)}$ among actions in $\Omega'$. Let the actions with rank smaller than $x_i^{(u)}$ be $L'$, then the p-value calculated from the sampled actions ($pv'$) = $\frac{|L'|}{|\Omega'|}$. Let $C = \{c_1, c_2 \ldots c_j \ldots c_N\}$ where $|\Omega| = N$ such that $c_j = 1$ if the action $a_j$ has a lower rank than $x_i^{(u)}$, and $c_j = 0$ otherwise. Then $C$ has $|L|$ ones and $N - |L|$ zeroes. Let $\{Y\}_{i=1}^n$ be the sampled actions where $Y_i$ represent the $i^{th}$ draw without replacement from $C$, such that $|\Omega'| = n$. Let $S_n = \sum_{i=1}^n Y_i$, which makes $S_n = |L'|$. Let $\bar{Y}_n = \frac{S_n}{n}$ which makes $\bar{Y}_n = \frac{|L'|}{|\Omega'|}$ and $\mu_{|L|,N} = \frac{|L|}{|\Omega|}$, then $S_n =: S_{n,|L|,N} \sim$ Hypergeometric$(n, |L|, N)$. Applying the Serfling bound [16]

$$P(\sqrt{n}(\bar{Y}_n - \mu_{|L|,N}) \geq \lambda) \leq exp\left(\frac{-2\lambda^2}{1 - \frac{n-1}{N}}\right),$$

By letting $\delta = pv'(x_i) - pv(x_i) = \bar{Y}_n - \mu_{|L|,N}$, let $\lambda = \epsilon\sqrt{|\Omega'|}$ yields

$$P(\delta \geq \epsilon) \leq exp\left(\frac{-2\epsilon^2|\Omega'|}{1 - \frac{|\Omega'|-1}{|\Omega|}}\right).$$

At $|\Omega| \gg |\Omega'|$, then: $P(\delta \geq \epsilon) \leq exp(-2\epsilon^2|\Omega'|)$. The bound no longer depends on $|\Omega|$, then we can fix $|\Omega'|$ to a constant.

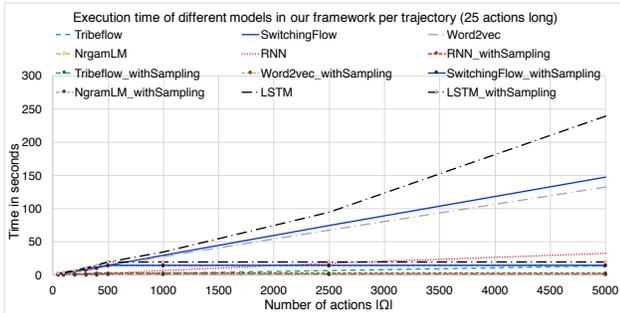

(a) The execution time for all models with and without sampling.

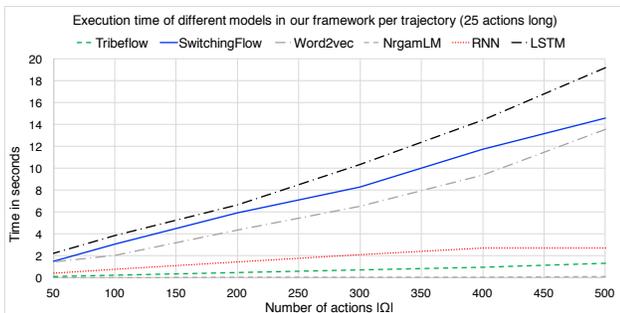

(b) The execution time for all models with sampling zoomed-in over the range $|\Omega'|$=50 to 500.

Figure 7: The execution time of different models in our framework using a single core machine for outlier detection of a 25 action long sequence from Last.Fm